\documentclass[runningheads]{llncs}

\usepackage{eccv}

\usepackage{eccvabbrv}

\usepackage{graphicx}
\usepackage{booktabs}

\usepackage[accsupp]{axessibility}  

\usepackage{amsmath}
\usepackage{amssymb}
\usepackage{amsfonts}
\usepackage{algorithmic}
\usepackage{graphicx}
\usepackage{multirow}
\usepackage{textcomp}
\usepackage{booktabs}
\usepackage{xcolor}
\usepackage{diagbox}
\usepackage{dsfont}
\usepackage{pifont}
\usepackage{tabularx}

\newcommand{\cmark}{\ding{51}}%
\newlength\savewidth\newcommand\shline{\noalign{\global\savewidth\arrayrulewidth\global\arrayrulewidth 1pt}\hline\noalign{\global\arrayrulewidth\savewidth}}

\usepackage[pagebackref,breaklinks,colorlinks,citecolor=eccvblue]{hyperref}

\usepackage{orcidlink}

\begin{document}

\title{NuNext: Reframing Nucleus Detection as Next-Point Detection} 





\author{
    Zhongyi Shui\inst{1,2,6}\thanks{Equal contribution} \and
    Honglin Li\inst{1,2}\textsuperscript{*} \and
    Xiaozhong Ji\inst{3} \and
    Ye Zhang\inst{4} \and
    Zijiang Yang\inst{5} \and
    Chenglu Zhu\inst{2} \and
    Yuxuan Sun\inst{1,2} \and
    Kai Yao\inst{1} \and
    Conghui He\inst{6} \and
    Cheng Tan\inst{6,\dag}
}

\authorrunning{Z. Shui et al.}

\institute{
  $^1$ Zhejiang University \quad
  $^2$ Westlake University \quad
  $^3$ Nanjing University \\
  $^4$ Harbin Institute of Technology \quad
  $^5$ University of Science and Technology Beijing \\
  $^6$ Shanghai Artificial Intelligence Laboratory \\
  \email{shuizhongyi@zju.edu.cn}\\
  \inst{\dag}Corresponding Author
}

\maketitle

\begin{abstract}
Nucleus detection in histopathology is pivotal for a wide range of clinical applications. Existing approaches either regress nuclear proxy maps that require complex post-processing, or employ dense anchors or queries that introduce severe foreground-background imbalance. In this work, we reformulate nucleus detection as next-point prediction, wherein a multimodal large language model is developed to directly output foreground nucleus centroids from the input image. The model is trained in two stages. In the supervised learning stage, we propose spatial-aware soft supervision to relax strict centroid matching and a chain-of-visual-thought strategy to incorporate visual priors that facilitate coordinate prediction. In the reinforcement fine-tuning stage, we design distribution matching reward, low-variance group filtering, and fine-grained advantage shaping to further improve the model's detection quality. Extensive experiments on nine widely used benchmarks demonstrate the superiority of our method. Code will be released soon.
\end{abstract}

\section{Introduction}
\label{sec:intro}

Nucleus detection is a fundamental task in computational pathology (CPath), underpinning a broad spectrum of downstream analyses of whole slide images (WSIs) such as cell counting \cite{ghahremani2022deep}, tumor microenvironment characterization \cite{rong2023deep} and spatial analysis of cellular organization \cite{jaume2024hest}. These analyse facilitate numerous clinically relevant applications including cancer subtyping \cite{chan2023histopathology}, grading \cite{amgad2022nucls}, staging \cite{yang2024histopathology}, prognosis assessment \cite{yang2024histopathology}, and treatment planning \cite{diao2021human}. Consequently, developing precise and generalizable nucleus detection algorithms has attracted increasing attention in the CPath community \cite{graham2019hover,horst2024cellvit,shui2024unleashing,zhang2025four}.

\begin{figure}[t]
  \centering
  \includegraphics[width=\linewidth]{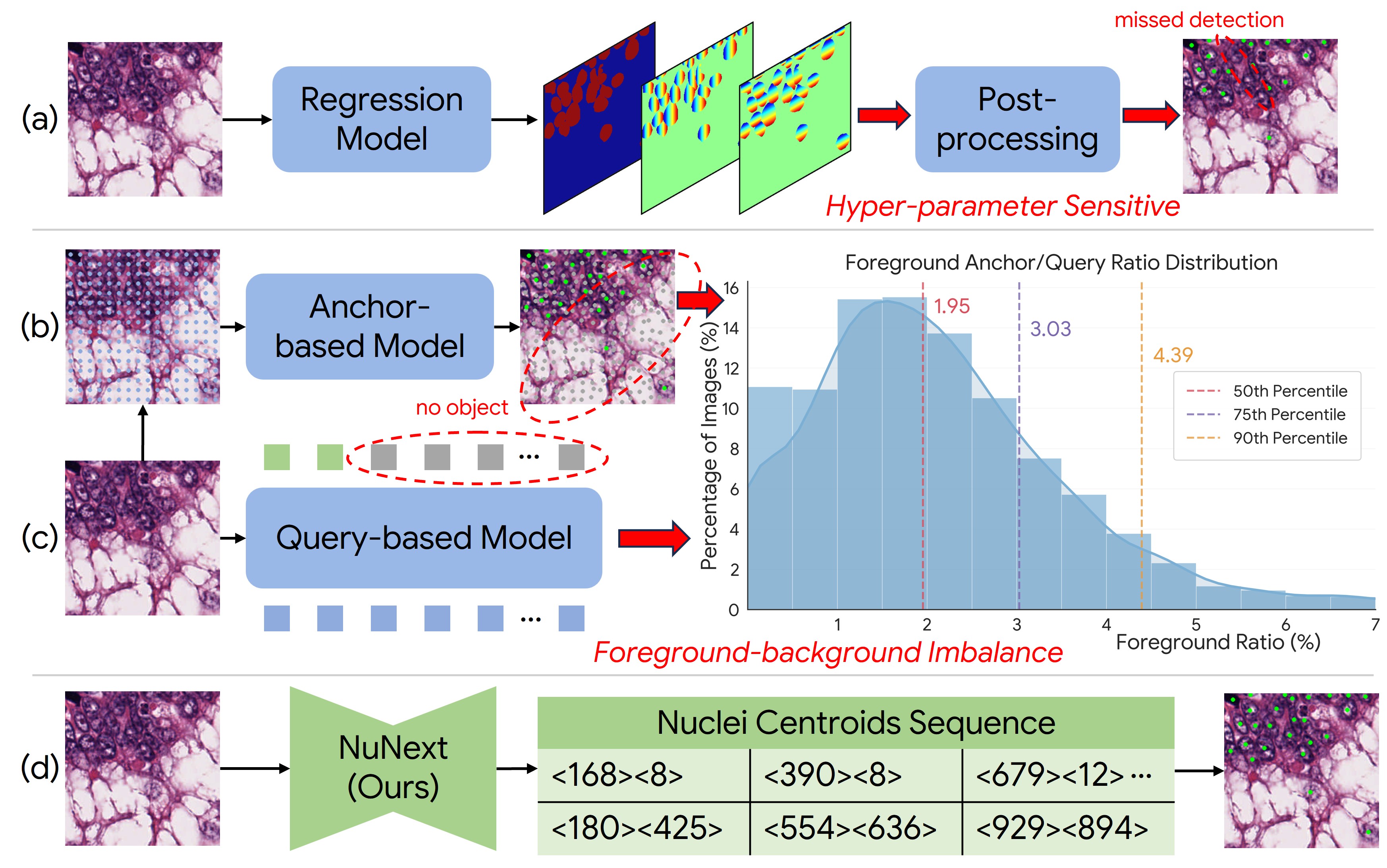}
  \caption{Pipeline comparison between nucleus detection paradigms. (a) Density map-based methods require complicated post-processing, which is hyper-parameter sensitive and vulnerable to noise. (b) Anchor-based and (c) query-based methods suffer from severe foreground-background imbalance, as the majority of anchors/queries are assigned to background. The histogram shows the foreground proportion across the large-scale PanNuke dataset, which is below 4.5\% for over 90\% of images. (d) Our proposed NuNext circumvents these issues by directly predicting nuclei coordinates.}
  \label{fig:banner}
  \vspace{-6pt}
\end{figure}

Mainstream nucleus detection approaches follow a bottom-up workflow \cite{graham2019hover,pan2023smile,yao2023pointnu,horst2024cellvit,stringer2025cellpose3,zhang2025four,sun2026disco}, as shown in Fig.~\ref{fig:banner} (a). They train a regression model to predict a nuclear probability map along with several auxiliary maps, which are then processed through hand-crafted post-processing to separate individual instances during inference. Despite significant progress, this paradigm suffers from substantial engineering overhead: the auxiliary maps rely on careful manual design with task-specific domain knowledge, and the inference pipelines involve numerous hand-tuned hyper-parameters that are highly sensitive to data characteristics.
To address these, \cite{song2021rethinking,shui2024dpa,yang2025muse} identify nuclei instances directly by refining and classifying predefined anchor points on the input image, and query-based methods \cite{huang2023affine,lou2025instance,pina2025cell} further replace predefined anchors with learnable object queries. However, both lines of methods rely on a large number of candidate anchors or queries to ensure adequate coverage in densely packed regions, which inevitably introduces substantial redundancy in sparse areas and leads to severe foreground-background imbalance, as illustrated in Fig.~\ref{fig:banner} (b) and (c).

In this work, we propose NuNext, which tokenizes continuous nuclear coordinates into discrete location tokens and reformulates nucleus detection as an autoregressive next-point prediction task (see Fig.~\ref{fig:banner} (d)). The model is trained in two stages. In the first stage, we perform supervised training with two key designs: a spatial-aware soft supervision strategy that relaxes strict centroid matching, and a chain-of-visual-thought technique that captures nucleus spatial information and provides visual priors for coordinate prediction. In the second stage, we transition from off-policy supervised learning to on-policy reinforcement fine-tuning, allowing the model to learn from its own sampling experience. Specifically, we adopt Group Relative Policy Optimization (GRPO) \cite{shao2024deepseekmath} for optimization, where a distribution matching reward is introduced to estimate detection quality. Additionally, we propose two improvements to GRPO: a low-variance group filtering strategy to suppress noisy gradient signals, and a fine-grained advantage shaping mechanism that enables token-level credit assignment. Finally, we adapt NuNext to nucleus instance segmentation by integrating it with PromptNucSeg \cite{shui2024unleashing}, and incorporate a task-guided reward into GRPO to improve the adaptation.

Our contributions can be summarized as follows:
\begin{itemize}
    \item We propose NuNext, which opens a new paradigm for nucleus detection through generative next-point prediction.
    \item We propose spatial-aware soft supervision and chain-of-visual-thought to boost coordinate prediction during supervised training.
    \item We tailor GRPO for nucleus detection with a distribution matching reward, and propose low-variance group filtering and fine-grained advantage shaping to further improve performance.
    \item Extensive experiments on nine challenging benchmarks demonstrate the effectiveness and superior generalization of our proposed method.
\end{itemize}

\section{Related Work}
\subsection{Nucleus Detection and Instance Segmentation}
Existing methods for nucleus detection and instance segmentation can be broadly categorized into three paradigms: density-map based, anchor-based, and query-based.

Density-map based methods first predict a nuclear probability map along with several auxiliary maps, and then group pixels into individual instances through meticulous post-processing. The design of auxiliary maps varies across methods. Specifically, CIA-Net \cite{zhou2019cia} and TSFD-Net \cite{ilyas2022tsfd} predict nuclear contour maps. DIST \cite{naylor2018segmentation}, StarDist \cite{schmidt2018cell}, HoVer-Net \cite{graham2019hover} and CellViT \cite{pina2025cell,horst2026cellvit++} design nuclear distance maps. The Cellpose series methods \cite{stringer2021cellpose,pachitariu2022cellpose,stringer2025cellpose3} predict gradient flow fields toward nucleus centers. FCIS \cite{zhang2025four} constructs four-color labeling maps, and Disco \cite{sun2026disco} adopts an adaptive coloring strategy that uses two colors for bipartite regions and a dedicated conflict color for topologically complex areas. Despite notable progress, these methods rely on hand-crafted post-processing pipelines that are parameter-sensitive and difficult to generalize, which presents a hurdle to their practical application.

To address this issue, anchor-based methods \cite{he2017mask,shui2024dpa,li2025nuhtc,shui2025towards} directly identify nucleus instances by refining and classifying predefined anchor points or boxes. However, the anchor layout still requires careful manual design, and anchors must be densely placed to cover tightly packed regions, inevitably leading to severe foreground-background imbalance in sparse areas. Although query-based methods such as AC-Former \cite{huang2023affine}, CellNuc-DETR \cite{pina2025cell}, and CellSAM \cite{marks2025cellsam} replace hand-crafted anchors with learnable object queries, they still require a large number of queries to avoid missing detections in crowded scenes, resulting in the same imbalance issue. Orthogonal to the above approaches, PromptNucSeg \cite{shui2024unleashing} decouples nucleus instance segmentation into two stages, first detecting nuclei centroids as point prompts, and then feeding them into SAM \cite{kirillov2023segment} to produce instance masks. This framework has demonstrated impressive performance in this field.

\subsection{Multimodal Generative Foundation Models in Pathology}
Current multimodal generative foundation models for computational pathology can be broadly categorized into three classes based on their input scale and reasoning scope: patch-level, slide-level, and omni-scale. At the patch level, Quilt-LLaVA \cite{seyfioglu2024quilt}, PathChat \cite{lu2024multimodal} and Patho-R1 \cite{zhang2025patho} adapt general-purpose Multimodal Large Language Models (MLLMs) to pathological tile interpretation via instruction tuning with domain-specific data. These models facilitate tile-level captioning, visual question answering (VQA) and interactive dialogue. At the slide level, WsiCaption \cite{chen2024wsicaption} and HistGen \cite{guo2024histgen} focus on generating slide-level reports from gigapixel WSIs, while SlideChat \cite{chen2025slidechat} and WSI-LLaVA \cite{liang2025wsi} extend the model's capability to slide-level VQA and multi-turn conversations. At the omni-scale level, CPath-Omni \cite{sun2025cpath} unifies a wide range of patch- and WSI-level tasks within a single framework. PathFinder \cite{Ghezloo_2025_ICCV} and CPathAgent \cite{sun2025cpathagent} emulate the diagnostic workflows of pathologists on WSIs by modeling multi-step navigation, providing more interpretable predictions.
 
Overall, existing works mainly focus on high-level semantic interpretation of pathological images, leaving the potential of MLLMs for dense prediction in pathology largely underexplored. To the best of our knowledge, NuNext is the first to leverage MLLMs for nucleus detection, broadening the frontier of multimodal foundation models in computational pathology from semantic understanding to fine-grained visual perception.

\begin{figure*}[t!]
  \centering
  \includegraphics[width=\linewidth]{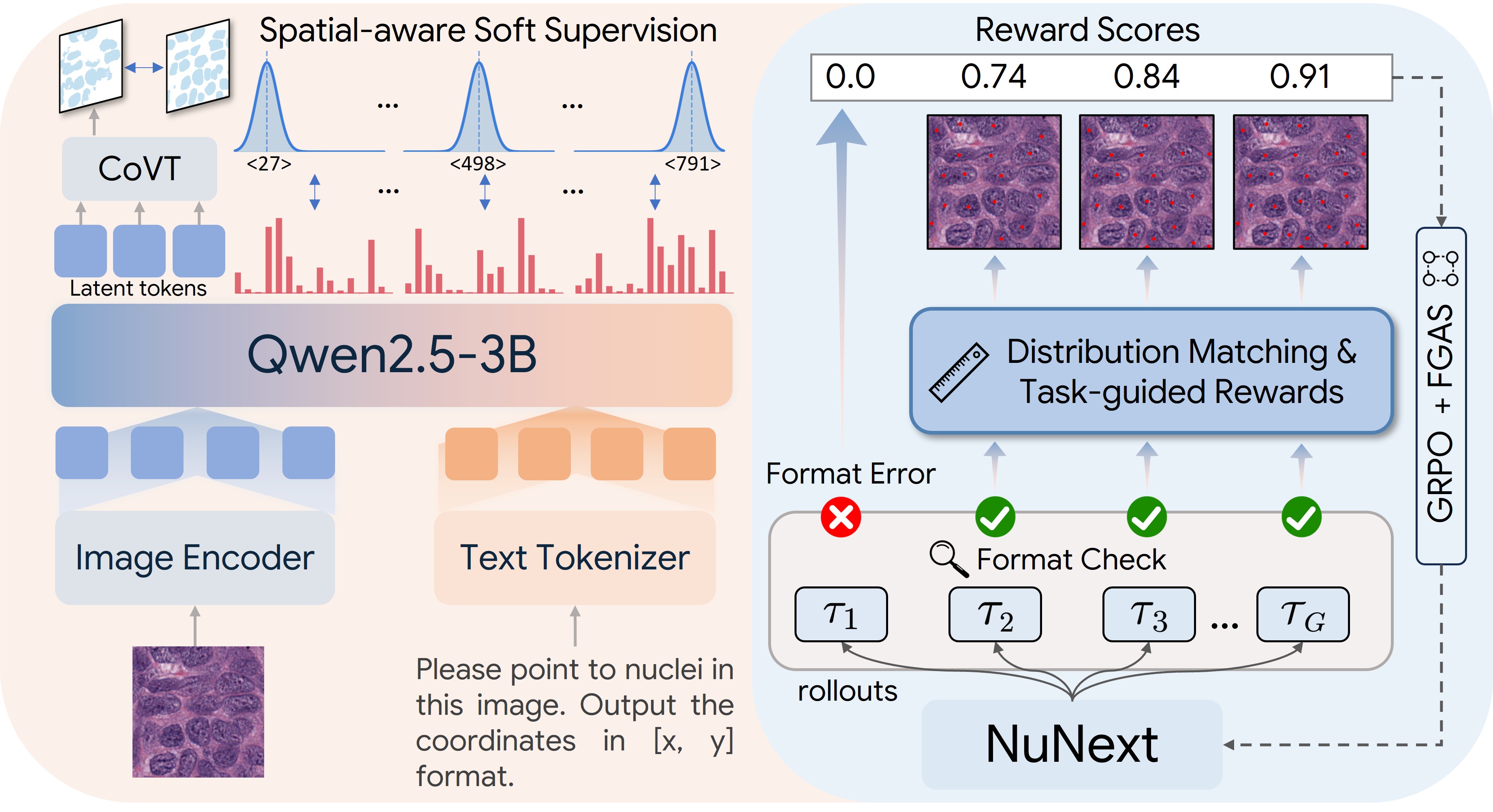}
  \caption{The training pipeline of NuNext. (Left) In the SFT stage, the model is trained to generate nucleus coordinate tokens with chain-of-visual-thought (CoVT) to incorporate visual cues of nuclei regions, and spatial-aware soft supervision that credits spatially proximate predictions. (Right) In the RFT stage, multiple rollouts are sampled per input, verified for format correctness, and scored with distribution matching and task-guided rewards. The model is then optimized via GRPO with fine-grained advantage shaping (FGAS).}
  \label{fig:method}
  \vspace{-5pt}
\end{figure*}

\section{Method}
\subsection{Overview}
In this work, we formulate nucleus detection as a \emph{next-point prediction} task, where nucleus centroids are represented as discrete coordinate tokens and generated autoregressively by an MLLM. The model is trained in two stages: supervised fine-tuning (SFT) followed by reinforcement fine-tuning (RFT). We adopt Qwen2.5-VL-3B \cite{bai2025qwen2,jiang2025detect} as our base model, and the overall training pipeline is depicted in Fig.~\ref{fig:method}.

\subsection{Coordinate Tokenization}
We represent continuous image coordinates with $K$ special coordinate tokens, which are appended to the original vocabulary of the language model.
Specifically, we uniformly quantize the normalized spatial interval $[0,1]$ into $K$ bins and assign each bin index $i\in\{0,\ldots,K-1\}$ to a unique token. The same token vocabulary is shared for both the $x$- and $y$-axes. Given a nucleus with coordinate $(x, y)$ in an image of width $W$ and height $H$, we first normalize it as $\tilde{x}=x/W$ and $\tilde{y}=y/H$ and then quantize each component to the nearest coordinate token:
\begin{equation}
\begin{aligned}
t^x &= \operatorname*{argmin}_{i\in\{0,\ldots,K-1\}}
\left|\tilde{x} - \frac{i+0.5}{K}\right|\\
t^y &= \operatorname*{argmin}_{i\in\{0,\ldots,K-1\}}
\left|\tilde{y} - \frac{i+0.5}{K}\right|
\end{aligned}
\end{equation}
This transforms continuous coordinate regression problem into a bounded $K$-way classification task. For an image containing $N$ nuclei, we apply this coordinate tokenization to each of them and concatenate the results into a token sequence $\mathcal{T} = \left(t_1^x, t_1^y, t_2^x, t_2^y, \ldots, t_N^x, t_N^y\right)$, where coordinate pairs are separated by a comma token. For notational convenience, we reindex the sequence as $\mathcal{T} = (t_1, t_2, \ldots, t_{2N})$ hereafter.

\subsection{Supervised Fine-tuning}
In the supervised fine-tuning stage, we train the model for end-to-end nucleus detection. Given a training sample $\mathcal{D}=\{(I, L, \mathcal{T})\}$, where $I$ denotes the input image, $L$ is the language instruction (\eg, ``Point to nucleus. Output the coordinates in [x, y] format.''), and $\mathcal{T}$ represents the nuclei coordinate token sequence. We encode the image $I$ into visual tokens $\mathcal{V}$ and the instruction $L$ into prompt tokens $\mathcal{Q}$, which are then concatenated and fed into a language model to predict the coordinate token sequence. The model is optimized via the standard next-token prediction (NTP) loss \cite{radford2018improving}:
\begin{equation}\label{eq:ntp}
\mathcal{L}_{\mathrm{NTP}}(\theta) = - \sum_{n=1}^{|\mathcal{T}|} \log P_\theta(t_n \mid \mathcal{V}, \mathcal{Q}, \mathcal{T}_{<n})
\end{equation}
in which $\theta$ denotes the model parameters, and $\mathcal{T}_{<n}$ represents the sequence of all preceding coordinate tokens. Notably, we sort coordinates in $\mathcal{T}$ by raster-scan order, \ie, from top to bottom and left to right. This strategy eliminates permutation ambiguity and provides consistent supervision for stable training.

\noindent\textbf{Spatial-Aware Soft Supervision.} While the next-token prediction objective in Eq.~\ref{eq:ntp} is widely employed in autoregressive generative tasks, we argue that it neglects the underlying spatial structure of images and is inherently suboptimal for nucleus detection. Specifically, at each decoding step, the model is supervised with a one-hot label to maximize the likelihood of the exact target coordinate token while treating all other tokens as incorrect, even those spatially proximal to the ground truth. This limitation is further amplified by the implicit weighting mechanism of the NTP loss. To illustrate, we analyze the gradient of the per-token loss with respect to the logits. Let $\mathbf{z}$ denote the logit vector produced by the language model, where $z_i$ is the raw score for the $i$-th token in the vocabulary and $p_i = \mathrm{softmax}(\mathbf{z})_i$ denotes its predicted probability. For simplicity, we consider only the $K$ coordinate tokens in the vocabulary here. For a decoding step with ground-truth coordinate token $t_n \in \{0, \ldots, K-1\}$, the loss in Eq.~\ref{eq:ntp} reduces to:
\begin{equation}
    \mathcal{L}_{\mathrm{NTP}} = -\log p_{t_n}
\end{equation}
Since $p_{t_n} = \frac{e^{z_{t_n}}}{\sum_{k=0}^{K-1} e^{z_k}}$, the loss can be rewritten as:
\begin{equation}
    \mathcal{L}_{\mathrm{NTP}} = -z_{t_n} + \log \sum_{k=0}^{K-1} e^{z_k}.
    \label{eq:ce_expand}
\end{equation}
Taking the derivative with respect to the logit $z_i$:
\begin{equation}
    \frac{\partial \mathcal{L}_{\mathrm{NTP}}}{\partial z_i} = -\mathds{1}{[i = t_n]} + \frac{e^{z_i}}{\sum_{k=0}^{K-1} e^{z_k}} = p_i - \mathds{1}{[i = t_n]},
    \label{eq:ce_deriv}
\end{equation}
where $\mathds{1}[i = t_n]$ is an indicator function that equals 1 if $i = t_n$ and 0 otherwise. By distinguishing the cases $i = t_n$ and $i \neq t_n$, we obtain:
\begin{equation}
    \frac{\partial \mathcal{L}_{\mathrm{NTP}}}{\partial z_i} =
    \begin{cases}
        p_i - 1, & i = t_n, \\
        p_i, & i \neq t_n.
    \end{cases}
    \label{eq:ce_gradient}
\end{equation}
With gradient descent optimization \cite{loshchilov2017decoupled}, this reveals two asymmetric behaviors: for the target token ($i = t_n$), the gradient is negative and its magnitude $1 - p_i$ decreases as the predicted probability increases, providing less encouragement once the model is already confident. In contrast, for non-target tokens ($i \neq t_n$), the gradient is positive with magnitude $p_i$, meaning that tokens with higher predicted probabilities receive stronger suppression. Due to the visual continuity of images, we empirically observed that the model tends to predict higher probabilities for tokens near the ground-truth coordinate than for distant ones in the early training stage. However, as analyzed above, with one-hot labels, higher predicted probabilities for non-target tokens lead to larger suppressive gradients, causing the model to penalize near-correct tokens more aggressively than distant ones. This hinders the model from exploiting the continuous nature of the coordinate space, making it easily trapped in local minima.

To mitigate this issue, we reformulate the supervision signal by replacing the one-hot label with a Gaussian-smoothed soft distribution that incorporates geometric relationships between coordinate tokens. For a ground-truth coordinate token $t_n \in \{0, \ldots, K-1\}$, the soft label over the $K$ coordinate bins is defined as:
\begin{equation}
s(k \mid t_n) = \frac{\exp\left(-\frac{(k - t_n)^2}{2\sigma^2}\right)}{\sum_{j=0}^{K-1} \exp\left(-\frac{(j - t_n)^2}{2\sigma^2}\right)},
\end{equation}
where $\sigma$ controls the smoothing strength. The training objective in Eq.~\ref{eq:ntp} then becomes:
\begin{equation}
\mathcal{L}_{\mathrm{NTP}}(\theta) = -\frac{1}{|\mathcal{T}|}\sum_{n=1}^{|\mathcal{T}|} \sum_{k=0}^{K-1} s(k \mid t_n) \log P_\theta(k \mid \mathcal{V}, \mathcal{Q}, \mathcal{T}_{<n})
\end{equation}

\noindent\textbf{Chain-of-Visual-Thought.}
Chain-of-Thought (CoT) has demonstrated remarkable success in enhancing the capabilities of large language models on complex textual tasks \cite{wei2022chain,kojima2022large,shao2024deepseekmath,song2025codedance,lin2026mmfinereason}. However, its potential for dense visual perception tasks remains largely unexplored. In this work, we propose a Chain-of-Visual-Thought (CoVT) strategy that introduces intermediate visual reasoning into the nucleus detection pipeline.

Specifically, we prepend a set of latent tokens $\mathcal{H} = \left(h_1, h_2, \ldots, h_L\right)$ to the original decoding sequence $\mathcal{T}$. These tokens are first generated by the language model, and then serve as prompt tokens for a frozen SAM \cite{kirillov2023segment} to predict binary nuclei foreground mask $\hat{M} = \mathrm{SAM}(I, \mathcal{H})$. We supervise this mask prediction with a combination of binary cross-entropy and dice loss.
\begin{equation}
\mathcal{L}_{\mathrm{CoVT}} = \mathcal{L}_{\mathrm{BCE}}\left(\hat{M},M\right) + \mathcal{L}_{\mathrm{Dice}}\left(\hat{M},M\right)
\end{equation}
in which $M$ denotes the ground-truth mask. Since SAM is pre-trained to segment objects at positions specified by its input prompts, we keep SAM frozen during training, such that minimizing $\mathcal{L}_{\mathrm{CoVT}}$ drives the latent tokens to capture nucleus location information. This provides visual priors that facilitate subsequent coordinate prediction. The overall loss used in the supervised fine-tuning stage is:
\begin{equation}
\mathcal{L}_{\mathrm{SFT}} = \mathcal{L}_{\mathrm{NTP}} + \alpha\mathcal{L}_{\mathrm{CoVT}}
\end{equation}
in which $\alpha$ is a weight term.

\subsection{Reinforcement Fine-tuning}
Despite the effectiveness of SFT, the model is trained off-policy where ground-truth tokens are provided at each step. This results in an exposure gap at inference time: the model autoregressively generates the coordinate sequence, where early prediction errors can accumulate and degrade subsequent outputs. To bridge this gap, we introduce a reinforcement fine-tuning (RFT) stage that optimizes the model on-policy from its own generated sequences. Since nucleus detection is naturally verifiable: the correctness of a detected nucleus can be validated by its distance to ground-truth annotations, we design a distribution matching reward function that evaluates the overall detection quality of each generated sequence, and optimize the model via Group Relative Policy Optimization (GRPO) \cite{shao2024deepseekmath}.

\noindent\textbf{Preliminaries on GRPO.} GRPO has demonstrated strong effectiveness across diverse tasks \cite{shao2024deepseekmath,jiang2025detect,song2025codedance}, it removes the need for a dedicated critic model in PPO \cite{schulman2017proximal} by directly computing advantages from reward comparison among $G$ sampled solutions. Formally, let $\pi_{\theta_{\text{old}}}$ and $\pi_{\theta}$ denote the policy model (parameterized by $\theta$) before and after the update, respectively. For a question $q$ sampled from a task dataset $\mathcal{D}$, a group of $G$ candidate solutions $\tau_i \sim \pi_{\theta_{\text{old}}}(\cdot \mid q)$ are sampled and evaluated with a reward function $r(\cdot)$.
Building on the clipped surrogate objective of PPO, we write the objective $\mathcal{J}$ in an empirical expectation form:
\begin{align}
\mathcal{J}_{\text{GRPO}}(\theta)
=& \mathbb{E}_{q \sim \mathcal{D}, \{\tau_i\}_{i=1}^G \sim \pi_{\theta_{\text{old}}}(\cdot \mid q)}
\Bigg[
\frac{1}{G} \sum_{i=1}^{G} \frac{1}{|\tau_i|} \sum_{n=1}^{|\tau_i|} \min\bigg(
\frac{\pi_\theta(\tau_{i,n} \mid q, \tau_{i,<n})}{\pi_{\theta_{\text{old}}}(\tau_{i,n} \mid q, \tau_{i,<n})} A_i,
\nonumber \\
&\text{clip}\!\left(
\frac{\pi_\theta(\tau_{i,n} \mid q, \tau_{i,<n})}{\pi_{\theta_{\text{old}}}(\tau_{i,n} \mid q, \tau_{i,<n})},
1\!-\!\epsilon, 1\!+\!\epsilon
\right) A_i
\bigg)
\Bigg]
\label{eq:grpo}
\end{align}
where $\epsilon = 0.2$ by default and $\text{clip}(\cdot)$ denotes the clipping operator for stability. We omit the Kullback–Leibler (KL) penalty for simplicity here. The normalized within-group reward then defines the advantage $A_i$ of solution $\tau_i$:
\begin{equation}
A_i
=
\frac{
r(\tau_i) - \operatorname{mean}\left( \{ r(\tau_j) \}_{j=1}^{G} \right)
}{
\operatorname{std}\left( \{ r(\tau_j) \}_{j=1}^{G} \right)
}
\label{eq:grpo_advantage}
\end{equation}

\noindent\textbf{Format check.} During SFT, the model is trained to output nuclei coordinates in a structured format: $(t_1^x, t_1^y, \ldots, t_N^x, t_N^y)$, establishing a parsing protocol for coordinate extraction. However, GRPO relies on high sampling temperature \cite{radford2018improving} to generate diverse rollouts, which often produces malformed outputs that lead to incorrect coordinate parsing and noisy reward signals. To deal with this, we verify the output format via regular expression matching and directly assign zero rewards to rollouts that fail the check.

\begin{figure*}[t!]
  \centering
  \includegraphics[width=0.99\linewidth]{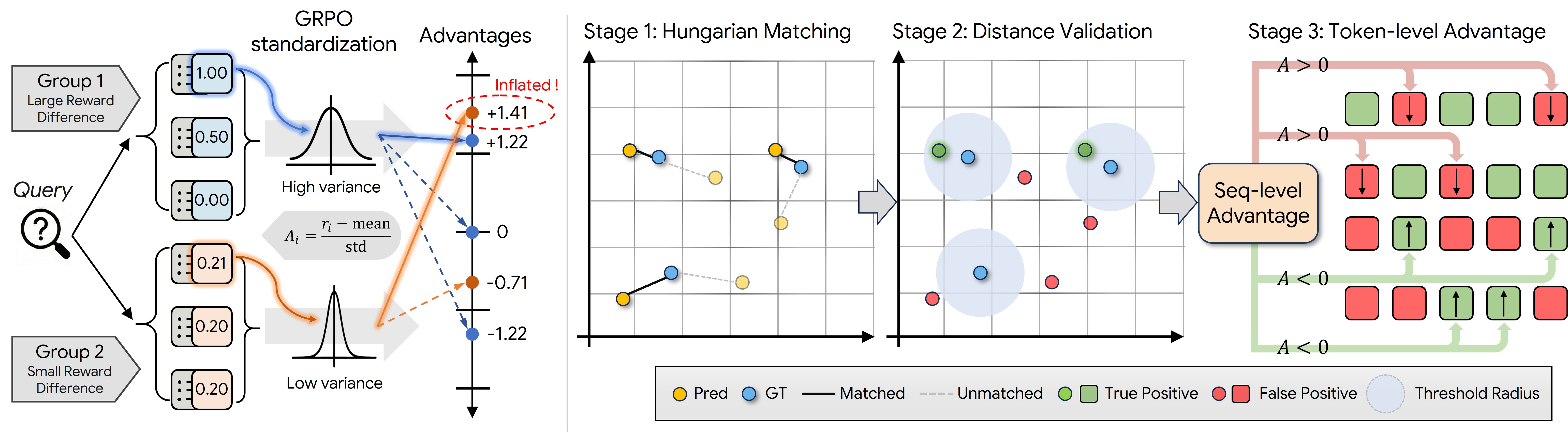}
  \caption{(Left) Motivation for low-variance group filtering: GRPO standardization can inflate advantages when within-group reward difference is negligible.
  (Right) Illustration of FGAS. Predicted and ground-truth nuclei are first matched via the Hungarian algorithm, then validated against a distance threshold to determine true/false positives. The resulting token-level labels are used to shape the sequence-level advantage, reducing the advantage for false positive tokens in rollouts with $A>0$ and alleviating the penalty for true positive tokens when $A<0$.} 
  \vspace{-5pt}
  \label{fig:motivation}
\end{figure*}

\noindent\textbf{Distribution matching reward.}
To evaluate the detection quality of each rollout, we introduce a distribution matching reward. Specifically, we first extract coordinate tokens from each rollout $\tau$ and convert them back to image coordinates. Let $\hat{\mathcal{C}}=\{\hat{c}_i\}_{i=1}^{\hat{N}}$ and $\mathcal{C}=\{c_j\}_{j=1}^{N}$ denote the predicted and ground-truth nucleus centroids, respectively, and $d_{ij}=\lVert \hat{c}_i - c_j\rVert_2$ be the Euclidean distance. Then, we establish a one-to-one matching between $\hat{\mathcal{C}}$ and $\mathcal{C}$ by solving the following minimum-cost bipartite matching problem with Hungarian algorithm \cite{kuhn1955hungarian}:
\begin{equation}
\pi^{*}
=
\arg\min_{\pi \in \Pi}
\sum_{i=1}^{\min(\hat{N},N)} d_{i,\pi(i)}
\label{eq:hungarian}
\end{equation}
where $\Pi$ denotes the set of all valid assignments. With a pre-defined distance threshold $r$, we count true positives (TP) as:
\begin{equation}
\mathrm{TP}
=
\sum_{i=1}^{\min(\hat{N},N)}
\mathbb{I}\!\left[d_{i,\pi^{*}(i)} \le r\right]
\label{eq:tp_def}
\end{equation}
The number of false positives (FP) and false negatives (FN) are derived as:
\begin{equation}
\mathrm{FP}=\hat{N}-\mathrm{TP}, \qquad
\mathrm{FN}=N-\mathrm{TP}
\end{equation}
We then compute recall and precision as follows:
\begin{equation}
\mathrm{Precision}=\frac{\mathrm{TP}}{\mathrm{TP}+\mathrm{FP}}, \qquad
\mathrm{Recall}=\frac{\mathrm{TP}}{\mathrm{TP}+\mathrm{FN}}
\end{equation}
Finally, we calculate F1-score as the distribution matching reward:
\begin{equation} r_{\text{DM}} = \frac{2 \cdot \mathrm{Precision} \cdot \mathrm{Recall}}{\mathrm{Precision} + \mathrm{Recall}}
\end{equation}

\noindent\textbf{Low-Variance Group Filtering.}
In this study, we identify a potential issue in GRPO: the advantage standardization can amplify negligible reward differences into strong gradient signals. To illustrate, assume a group size of 3. For the same input $q$, consider two groups $\mathcal{G}_1$ and $\mathcal{G}_2$ with rewards $\{0, 0.5, 1.0\}$ and $\{0.20, 0.20, 0.21\}$, respectively. After GRPO standardization, the rollout with $r=0.21$ in $\mathcal{G}_2$ receives the largest advantage across both groups, surpassing even those with higher rewards in $\mathcal{G}_1$, as illustrated in Fig.~\ref{fig:motivation} (Left). This is because GRPO computes advantages through reward standardization, \ie, subtracting the mean and dividing by the standard deviation, as shown in Eq.~\ref{eq:grpo_advantage}. When rewards within a group are nearly identical, the small standard deviation amplifies negligible differences into disproportionately strong gradient signals, introducing noise into training. To address this, we dynamically filter out groups with reward standard deviation below a threshold $\delta$ during training.

\noindent\textbf{Fine-grained Advantage Shaping.}
Common RL methods such as GRPO \cite{shao2024deepseekmath} and DAPO \cite{yu2025dapo} rely on sequence-level rewards, where all tokens within a generated rollout share the same advantage, as shown in Eq.~\ref{eq:grpo}. This leads to a credit assignment issue \cite{sutton1998reinforcement}: all tokens are equally encouraged or penalized regardless of their individual contribution to the final reward. In this work, we observe that unlike text tokens whose contribution is context-dependent and difficult to assess, each coordinate token in our task corresponds to an explicit spatial location and can be directly evaluated against the ground-truth annotations. Based on the matching result in Eq.~\ref{eq:tp_def}, each predicted coordinate can be categorized as either a true positive or a false positive. Intuitively, in a rollout with advantage $A > 0$, tokens corresponding to false positive coordinates should not receive the same encouragement as true positives, and conversely, in a rollout with advantage $A < 0$, true positive tokens should not be penalized as heavily as false positives. Motivated by this, we propose a fine-grained advantage shaping (FGAS) strategy to enable token-level credit assignment:
\begin{equation}
\tilde{A}_{i,n} =
\begin{cases}
\beta \cdot A_i, & \text{if } (A_i > 0 \wedge \hat{c}_n \in \mathrm{FP}) \lor (A_i < 0 \wedge \hat{c}_n \in \mathrm{TP}), \\
A_i, & \text{otherwise}
\end{cases}
\end{equation}
where $\beta \in (0, 1)$ is a decay factor. We then replace $A_i$ with $\tilde{A}_{i,n}$ in the GRPO objective. The computation process of FGAS is illustrated in Fig.~\ref{fig:motivation} (Right).

\section{Adaption to Nucleus Instance Segmentation}
Our model can be naturally extended to nucleus instance segmentation by integrating with PromptNucSeg \cite{shui2024unleashing}, a two-stage pipeline where a detection model generates a point prompt for each nucleus and SAM then produces the corresponding instance mask. Although this pipeline has demonstrated strong performance on multiple benchmarks, we empirically observe that better detection does not necessarily lead to better instance segmentation. This is because detection only requires locating each nucleus within a certain distance threshold, whereas segmentation quality is affected by the prompt position within the nucleus. Furthermore, in the original PromptNucSeg, the detection model is optimized independently, receiving no feedback from downstream segmentation performance.

To narrow this gap, for each rollout $\tau$, we feed the extracted nuclei coordinates as point prompts into SAM to generate instance masks, and compute the instance segmentation metric Panoptic Quality (PQ) as an auxiliary task-guided reward. The overall reward used in the reinforcement fine-tuning stage is:
\begin{equation}
r_{\mathrm{RFT}} = r_{\mathrm{DM}} + \gamma \, r_{\mathrm{PQ}}
\end{equation}
in which $\gamma$ balances the two reward terms. Since PQ is non-differentiable, by incorporating it as a reward signal, RL enables direct optimization of the detection model toward end-task performance. Moreover, we observe that point prompts closer to the nucleus center generally yield higher-quality masks and thus higher PQ rewards. This in turn provides dense supervision that encourages more precise localization for the detection model, creating a synergy between detection and segmentation.

\section{Experiment}
\subsection{Experimental Settings}
\noindent\textbf{Datasets.} We train our model on the PanNuke dataset \cite{gamper2019pannuke,gamper2020pannuke}, the largest pan-cancer dataset for nuclei instance segmentation and classification. It contains 7,901 hematoxylin and eosin (H\&E) stained images of size 256$\times$256, covering 19 tissue types sampled from over 20,000 WSIs in TCGA \cite{weinstein2013cancer}. The dataset provides detailed annotations for 189,744 nuclei categorized into five categories, and is divided into three folds. In this work, we utilize Fold 1 and Fold 2 for training and Fold 3 for internal validation, following \cite{jo2025coin}. For fair comparison, all baseline models are trained using the same data split. For external validation, we evaluate all methods on eight benchmarks across various tissue types and acquisition conditions: CPM-15 \cite{vu2019methods}, CPM-17 \cite{vu2019methods}, Cryonuseg \cite{mahbod2021cryonuseg}, TNBC \cite{naylor2018segmentation}, BRCA-M2C \cite{abousamra2021multi}, Kumar \cite{kumar2019multi,kumar2017dataset}, GLySAC \cite{doan2022sonnet} and ConSeP \cite{graham2019hover}. A comprehensive introduction to these datasets is provided in the supplementary material.

\noindent\textbf{Evaluation metrics.} For nucleus detection, we employ F1-score as the evaluation metric. For nucleus instance segmentation, following previous works \cite{graham2019hover,yao2023pointnu,horst2024cellvit}, we adopt Aggregate Jaccard Index (AJI) and Panoptic Quality (PQ) as the evaluation metrics. PQ can be further decomposed into Detection Quality (DQ) and Segmentation Quality (SQ), providing more fine-grained evaluation.

\noindent\textbf{Implementation details} are available in the appendix.

\begin{table*}[t!]
	\centering
    \caption{Performance comparison on PanNuke. The best and second-best PQ scores are highlighted in \textbf{bold} and \underline{underlined}.}
	\small{
		\resizebox{0.99\linewidth}{!}{		
			\begin{tabular}{c|cc|cc|cc|cc|cc|cc|cc}
				\toprule[1.5pt]
				\multirow{3}{*}{Tissue} & \multicolumn{2}{c|}{Hover-Net} & \multicolumn{2}{c|}{CPP-Net} & \multicolumn{2}{c|}{PointNu-Net} & \multicolumn{2}{c|}{CellViT-H} & \multicolumn{2}{c|}{DPA-P2PNet} & \multicolumn{2}{c|}{Cell-DETR} & \multicolumn{2}{c}{NuNext} \\
                & \multicolumn{2}{c|}{\cite{graham2019hover}} & \multicolumn{2}{c|}{\cite{chen2023cpp}} & \multicolumn{2}{c|}{\cite{yao2023pointnu}} & \multicolumn{2}{c|}{\cite{horst2024cellvit}} & \multicolumn{2}{c|}{\cite{shui2024dpa,shui2024unleashing}} & \multicolumn{2}{c|}{\cite{pina2025cell}} & \multicolumn{2}{c}{(Ours)} \\
				\cmidrule(lr){2-3} \cmidrule(lr){4-5} \cmidrule(lr){6-7} \cmidrule(lr){8-9} \cmidrule(lr){10-11} \cmidrule(lr){12-13} \cmidrule(lr){14-15}
				& bPQ & mPQ & bPQ & mPQ & bPQ & mPQ & bPQ & mPQ & bPQ & mPQ & bPQ & mPQ & bPQ & mPQ \\
				\shline
                Adrenal & 0.7007 & 0.4906 & 0.6932 & 0.4984 & 0.7219 & 0.5154 & 0.7105 & 0.5173 & 0.7300 & 0.5171 & \underline{0.7301} & \underline{0.5428} & \textbf{0.7417} & \textbf{0.5464} \\
                Bile Duct & 0.6584 & 0.4497 & 0.6573 & 0.4493 & 0.6672 & 0.4627 & 0.6733 & 0.4777 & \textbf{0.6970} & \textbf{0.4908} & 0.6933 & 0.4805 & \underline{0.6953} & \underline{0.4897} \\
                Bladder & 0.7068 & 0.5722 & 0.7030 & 0.5820 & \underline{0.7370} & \textbf{0.6233} & 0.7069 & 0.5683 & 0.7289 & 0.5988 & 0.7335 & \underline{0.6212} & \textbf{0.7505} & 0.6201 \\
                Breast & 0.6595 & 0.5177 & 0.6686 & 0.5138 & 0.6688 & 0.5139 & 0.6775 & 0.5279 & \underline{0.6874} & \textbf{0.5464} & 0.6831 & 0.5369 & \textbf{0.6925} & \underline{0.5432} \\
                Cervix & 0.6714 & 0.4743 & 0.6849 & 0.5133 & 0.6858 & 0.5126 & 0.6977 & \underline{0.5202} & 0.7022 & 0.4924 & \underline{0.7065} & 0.5168 & \textbf{0.7067} & \textbf{0.5318} \\
                Colon & 0.5713 & 0.4243 & 0.5762 & 0.4356 & 0.5900 & 0.4594 & 0.6055 & 0.4606 & 0.6110 & 0.4771 & \underline{0.6138} & \underline{0.4792} & \textbf{0.6359} & \textbf{0.4969} \\
                Esophagus & 0.6625 & 0.5288 & 0.6701 & 0.5423 & 0.6784 & 0.5534 & 0.6761 & 0.5474 & 0.6906 & \underline{0.5763} & \underline{0.7000} & 0.5720 & \textbf{0.7129} & \textbf{0.5876} \\
                Head \& Neck & 0.6078 & 0.4460 & 0.6396 & 0.4683 & 0.6451 & 0.4743 & 0.6499 & 0.4725 & 0.6580 & \underline{0.4903} & \underline{0.6593} & 0.4876 & \textbf{0.6859} & \textbf{0.5083} \\
                Kidney & 0.6826 & 0.5237 & 0.6997 & 0.5374 & 0.6986 & 0.5466 & \underline{0.7224} & 0.5711 & 0.7155 & 0.5711 & 0.7183 & \underline{0.5756} & \textbf{0.7277} & \textbf{0.5911} \\
                Liver & 0.7064 & 0.4933 & 0.7041 & 0.5054 & 0.7234 & 0.5269 & 0.7271 & 0.5229 & \underline{0.7385} & 0.5291 & 0.7283 & \textbf{0.5460} & \textbf{0.7473} & \underline{0.5425} \\
                Lung & 0.6307 & 0.3933 & 0.6379 & 0.4320 & 0.6504 & 0.4295 & 0.6439 & 0.4375 & \underline{0.6705} & \underline{0.4454} & 0.6617 & 0.4389 & \textbf{0.6813} & \textbf{0.4654} \\
                Ovarian & 0.6736 & 0.5358 & 0.6774 & 0.5536 & 0.6918 & 0.5543 & 0.6793 & \underline{0.5566} & \underline{0.7035} & 0.5469 & 0.6964 & \underline{0.5739} & \textbf{0.7077} & \textbf{0.5805} \\
                Pancreatic & 0.6562 & 0.4914 & 0.6707 & 0.4812 & 0.6742 & \underline{0.5161} & 0.6846 & 0.5052 & 0.6814 & 0.4973 & \textbf{0.7006} & \textbf{0.5266} & \underline{0.7005} & \underline{0.5254} \\
                Prostate & 0.6826 & 0.5368 & 0.6770 & 0.5363 & 0.6890 & 0.5290 & \underline{0.7046} & 0.5510 & 0.7025 & 0.5398 & 0.7018 & \underline{0.5563} & \textbf{0.7154} & \textbf{0.5719} \\
                Skin & 0.6328 & 0.3934 & 0.6285 & 0.4276 & 0.6293 & 0.4273 & \underline{0.6589} & \underline{0.4409} & \textbf{0.6648} & \textbf{0.4471} & 0.6459 & 0.4416 & 0.6555 & 0.4292 \\
                Stomach & 0.7047 & 0.4830 & 0.6932 & 0.4516 & 0.7171 & 0.4818 & 0.7218 & \underline{0.5007} & \underline{0.7259} & 0.4759 & 0.7150 & 0.4606 & \textbf{0.7314} & \textbf{0.5049} \\
                Testis & 0.7028 & 0.5307 & 0.6818 & 0.4987 & 0.6983 & \underline{0.5505} & 0.6937 & 0.5364 & \underline{0.7136} & \textbf{0.5525} & 0.7100 & 0.5477 & \textbf{0.7230} & 0.5447 \\
                Thyroid & 0.7128 & 0.4174 & 0.7038 & 0.4407 & 0.7119 & \underline{0.4699} & 0.7156 & 0.4615 & 0.7385 & 0.4746 & \underline{0.7407} & \underline{0.4866} & \textbf{0.7600} & \textbf{0.4911} \\
                Uterus & 0.6603 & 0.4647 & 0.6657 & 0.4546 & 0.6697 & 0.4585 & \underline{0.6760} & 0.4666 & \underline{0.6794} & \textbf{0.4958} & 0.6678 & \underline{0.4654} & \textbf{0.6941} & \underline{0.4886} \\
				\shline
                Average & 0.6676 & 0.4825 & 0.6701 & 0.4906 & 0.6815 & 0.5055 & 0.6856 & 0.5075 & \underline{0.6968} & 0.5139 & 0.6951 & \underline{0.5187} & \textbf{0.7087} & \textbf{0.5294} \\
				\bottomrule[1.5pt]
			\end{tabular}}}\label{tab:pannuke_seg}
            \vspace{-5pt}
\end{table*}

\begin{table}[t!]
	\centering
    \caption{Comparison of detection and instance segmentation performance across different nuclear categories on the PanNuke dataset.}
	\resizebox{0.99\linewidth}{!}{		
		\begin{tabular}{c|cc|cc|cc|cc|cc}
			\shline
			\diagbox[width=9em]{Method}{Class} 
			& \multicolumn{2}{c|}{Neoplastic} & \multicolumn{2}{c|}{Epithelial} & \multicolumn{2}{c|}{Inflammatory} & \multicolumn{2}{c|}{Connective} & \multicolumn{2}{c}{Necrosis} \\
			\cline{2-11}
			& F1 & PQ & F1 & PQ & F1 & PQ & F1 & PQ & F1 & PQ \\
			\shline
			Hover-Net \cite{graham2019hover} & 0.680 & 0.564 & 0.691 & 0.550 & 0.570 & 0.432 & 0.519 & 0.407 & 0.300 & 0.097 \\
            StarDist \cite{graham2019hover} & 0.706 & 0.556 & 0.740 & 0.562 & 0.580 & 0.395 & 0.525 & 0.362 & 0.339 & 0.115 \\
            CPP-Net \cite{graham2019hover} & 0.722 & 0.591 & 0.761 & 0.584 & 0.595 & 0.404 & 0.553 & 0.408 & 0.392 & 0.139 \\
            PointNu-Net \cite{yao2023pointnu} & 0.715 & 0.592 & 0.753 & 0.587 & 0.585 & 0.427 & 0.543 & 0.412 & 0.378 & 0.138 \\
			SMILE \cite{pan2023smile} & 0.667 & 0.536 & 0.667 & 0.496 & 0.567 & 0.425 & 0.499 & 0.384 & 0.293 & 0.088 \\
            CellViT-H \cite{horst2024cellvit}& 0.723 & 0.595 & 0.765 & 0.596 & 0.593 & 0.422 & 0.547 & 0.428 & 0.380 & 0.144 \\
            DPA-P2PNet \cite{shui2024dpa,shui2024unleashing} & \underline{0.726} & \underline{0.609} & 0.755 & 0.593 & 0.600 & 0.443 & 0.564 & \underline{0.444} & \textbf{0.416} & \underline{0.169} \\
			FCIS \cite{zhang2025four} & 0.681 & 0.560 & 0.716 & 0.547 & 0.580 & 0.437 & 0.544 & 0.414 & 0.386 & 0.128 \\
            CellNuc-DETR \cite{li2025nuhtc} & 0.725 & 0.602 & \underline{0.768} & \underline{0.607} & \underline{0.603} & \underline{0.446} & \underline{0.568} & 0.436 & 0.396 & \textbf{0.190} \\
			NuNext (Ours) & \textbf{0.733} & \textbf{0.617} & \textbf{0.775} & \textbf{0.615} & \textbf{0.604} & \textbf{0.454} & \textbf{0.577} & \textbf{0.449} & \underline{0.408} & 0.159 \\
			\shline
	\end{tabular}}\label{tab:pannuke_nuc}
    \vspace{-5pt}
\end{table}

\begin{figure*}[t!]
  \centering
  \includegraphics[width=\linewidth]{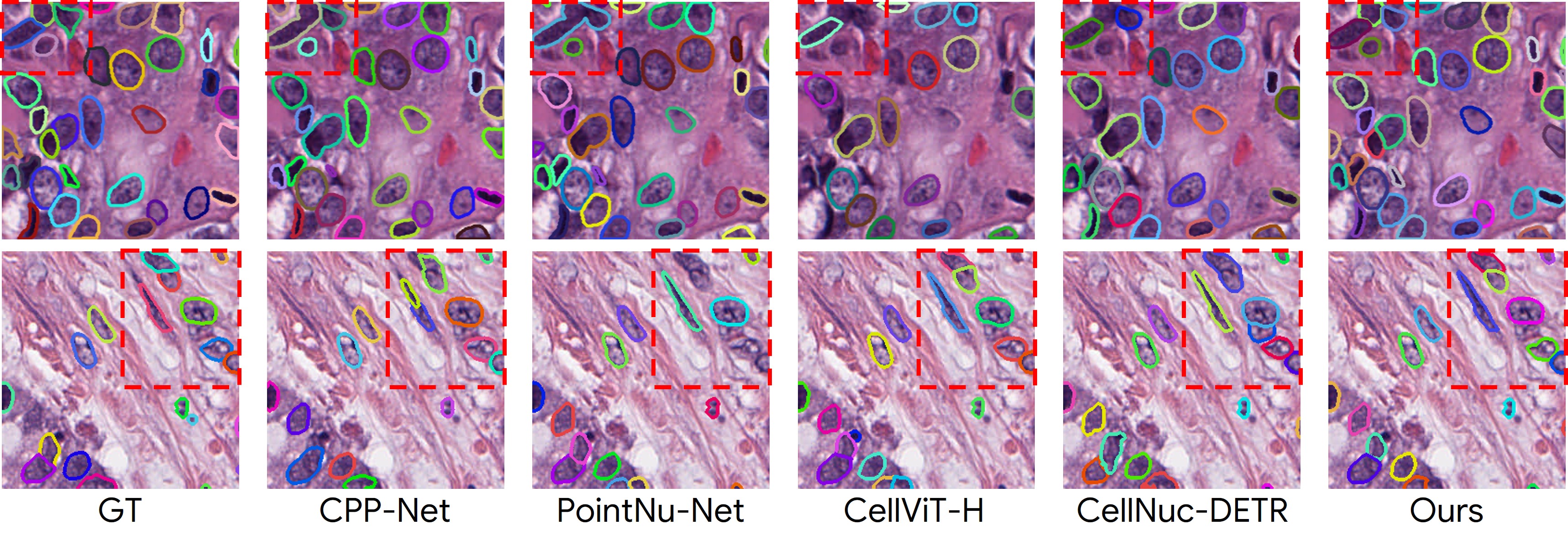}
  \vspace{-15pt}
  \caption{Qualitative comparison with SOTA methods.  
  }\label{fig:cmp}
  \vspace{-5pt}
\end{figure*}

\begin{table}[t!]
	\centering
	\caption{Generalization comparison on CPM-15, CPM-17, CryoNuSeg, and TNBC. The highest AJI and PQ scores are in \textbf{bold} while the second highest are \underline{underlined}.}
	\resizebox{0.99\linewidth}{!}{
		\begin{tabular}{c|cccc|cccc|cccc|cccc}
			\shline
			\diagbox[width=9em]{Method}{Dataset}
			& \multicolumn{4}{c|}{CPM-15}
			& \multicolumn{4}{c|}{CPM-17} 
            & \multicolumn{4}{c|}{CryoNuSeg}
            & \multicolumn{4}{c}{TNBC} \\
			\cline{2-17}
			& AJI & DQ & SQ & PQ & AJI & DQ & SQ & PQ & AJI & DQ & SQ & PQ & AJI & DQ & SQ & PQ \\
			\shline
			Hover-Net \cite{graham2019hover} & 0.631 & 0.764 & 0.820 & 0.629 & 0.715 & 0.864 & 0.812 & 0.703 & 0.459 & 0.597 & 0.763 & 0.458 & 0.652 & 0.813 & 0.803 & 0.653 \\
            StarDist \cite{graham2019hover} & 0.588 & 0.745 & 0.768 & 0.577 & 0.658 & 0.842 & 0.770 & 0.650 & 0.514 & 0.653 & 0.746 & 0.490 & 0.654 & 0.829 & 0.774 & 0.643 \\
            CPP-Net \cite{graham2019hover} & 0.601 & 0.757 & 0.778 & 0.593 & 0.674 & 0.856 & 0.785 & 0.673 & 0.517 & 0.650 & 0.757 & 0.494 & 0.665 & 0.844 & 0.782 & 0.663 \\
            PointNu-Net \cite{yao2023pointnu} & 0.601 & 0.742 & 0.821 & 0.612 & 0.673 & 0.848 & 0.818 & 0.694 & 0.434 & 0.562 & 0.775 & 0.437 & 0.652 & 0.820 & 0.785 & 0.658 \\
			SMILE \cite{pan2023smile} & 0.629 & 0.760 & 0.818 & 0.625 & \underline{0.718} & 0.866 & 0.812 & \underline{0.704} & \underline{0.524} & 0.655 & 0.767 & 0.505 & 0.649 & 0.815 & 0.795 & 0.650 \\
			CellViT-H \cite{horst2024cellvit} & 0.616 & 0.766 & 0.790 & 0.608 & 0.690 & 0.860 & 0.795 & 0.684 & \textbf{0.525} & 0.659 & 0.768 & \underline{0.509} & 0.666 & 0.834 & 0.795 & 0.665 \\
            DPA-P2PNet \cite{shui2024dpa,shui2024unleashing} & 0.520 & 0.676 & 0.819 & 0.556 & 0.590 & 0.775 & 0.826 & 0.640 & 0.515 & 0.655 & 0.770 & 0.507 & 0.671 & 0.839 & 0.797 & 0.669 \\
			FICS \cite{zhang2025four} & \underline{0.640} & 0.777 & 0.788 & 0.616 & 0.701 & 0.848 & 0.811 & 0.689 & 0.481 & 0.601 & 0.764 & 0.461 & \underline{0.688} & 0.835 & 0.799 & 0.668 \\
            CellNuc-DETR \cite{pina2025cell} & 0.616 & 0.771 & 0.815 & \underline{0.630} & 0.689 & 0.854 & 0.820 & 0.701 & 0.499 & 0.634 & 0.774 & 0.493 & 0.682 & 0.858 & 0.798 & \textbf{0.685} \\
			NuNext (Ours) & \textbf{0.656} & 0.798 & 0.816 & \textbf{0.653} & \textbf{0.723} & 0.881 & 0.821 & \textbf{0.724} & 0.523 & 0.675 & 0.769 & \textbf{0.521} & \textbf{0.698} & 0.863 & 0.792 & \underline{0.683} \\
			\shline
		\end{tabular}
	}
	\label{tab:generalization1}
    \vspace{-5pt}
\end{table}

\begin{table}[t!]
	\centering
	\caption{Generalization comparison on BRCA-M2C, Kumar, GLySAC and CoNSeP.}
	\resizebox{0.99\linewidth}{!}{
		\begin{tabular}{c|cccc|cccc|cccc|cccc}
			\shline
			\diagbox[width=9em]{Method}{Dataset}
			& \multicolumn{4}{c|}{BRCA-M2C}
			& \multicolumn{4}{c|}{Kumar} 
			& \multicolumn{4}{c|}{GLySAC}
			& \multicolumn{4}{c}{CoNSeP} \\
			\cline{2-17}
			& AJI & DQ & SQ & PQ & AJI & DQ & SQ & PQ & AJI & DQ & SQ & PQ & AJI & DQ & SQ & PQ \\
			\shline
			Hover-Net \cite{graham2019hover} & 0.702 & 0.787 & 0.877 & 0.692 & 0.642 & 0.80 & 0.786 & 0.630 & 0.586 & 0.746 & 0.744 & 0.560 & 0.533 & 0.670 & 0.766 & 0.515 \\
			StarDist \cite{schmidt2018cell} & 0.680 & 0.805 & 0.849 & 0.685 & 0.623 & 0.802 & 0.758 & 0.609 & 0.570 & 0.743 & 0.720 & 0.539 & 0.478 & 0.601 & 0.715 & 0.433 \\
			CPP-Net \cite{chen2023cpp} & 0.706 & 0.826 & 0.870 & 0.720 & 0.637 & 0.818 & 0.767 & 0.629 & 0.577 & 0.752 & 0.725 & 0.550 & 0.524 & 0.664 & 0.726 & 0.485 \\
			PointNu-Net \cite{yao2023pointnu} & 0.692 & 0.816 & 0.883 & 0.721 & 0.631 & 0.816 & 0.790 & 0.646 & 0.552 & 0.727 & 0.752 & 0.552 & 0.504 & 0.688 & 0.774 & 0.533 \\
			SMILE \cite{pan2023smile} & 0.708 & 0.799 & 0.885 & 0.709 & \underline{0.651} & 0.811 & 0.783 & 0.636 & \underline{0.594} & 0.752 & 0.743 & 0.563 & 0.444 & 0.572 & 0.746 & 0.429 \\
			CellViT-H \cite{horst2024cellvit} & 0.708 & 0.816 & 0.885 & \underline{0.723} & 0.644 & 0.820 & 0.771 & 0.633 & 0.572 & 0.736 & 0.728 & 0.541  & 0.560 & 0.703 & 0.743 & 0.525 \\
			DPA-P2PNet \cite{shui2024dpa,shui2024unleashing} & 0.671 & 0.788 & 0.893 & 0.704 & 0.585 &  0.772 & 0.795 & 0.615 & 0.508 & 0.682 & 0.750 & 0.517  & 0.521 & 0.686 & 0.779 & 0.537 \\
			FICS \cite{zhang2025four} & \textbf{0.714} & 0.806 & 0.885 & 0.714 & 0.622 & 0.780 & 0.786 & 0.614 & 0.587 & 0.743 & 0.759 & \underline{0.568} & 0.492 & 0.650 & 0.772 & 0.504 \\
            CellNuc-DETR \cite{pina2025cell} & 0.706 & 0.814 & 0.887 & \underline{0.723} & 0.641 & 0.821 & 0.789 & \underline{0.649} &  0.566 & 0.735 & 0.744 & 0.552 & \underline{0.577} & 0.730 & 0.768 & \underline{0.562} \\
			NuNext (Ours) & \underline{0.713} & 0.819 & 0.890 & \textbf{0.730} & \textbf{0.654} & 0.829 & 0.789 & \textbf{0.655} & \textbf{0.599} & 0.761 & 0.755 & \textbf{0.581} & \textbf{0.582} & 0.741 & 0.790 & \textbf{0.586} \\
			\shline
		\end{tabular}
	}
    \vspace{-5pt}
	\label{tab:generalization2}
\end{table}

\subsection{Comparison with SOTA methods}
We compare with state-of-the-art (SOTA) counterparts spanning three categories: density-map based approaches (\ie, CellViT-H \cite{horst2024cellvit,horst2026cellvit++}), anchor-based method (DPA-P2PNet \cite{shui2024dpa,shui2024unleashing}), and query-based method (CellNuc-DETR \cite{pina2025cell}). For methods that only output nuclei centroid coordinates, we adopt the same PromptNucSeg pipeline as ours to evaluate their instance segmentation performance. Tab.~\ref{tab:pannuke_seg} shows the quantitative comparison on the PanNuke benchmark. Without additional techniques such as test-time augmentation \cite{yao2023pointnu}, stain normalization \cite{pina2025cell}, class-balanced sampling or adding an auxiliary tissue classification branch \cite{horst2024cellvit,horst2026cellvit++}, our proposed model outperforms the best previous models by 1.19 bPQ and 1.07 mPQ. Moreover, we report the per-category detection and segmentation performance in Tab.~\ref{tab:pannuke_nuc}. Overall, NuNext achieves the best performance across four out of five nuclear categories.  Tab.~\ref{tab:generalization1} and Tab.~\ref{tab:generalization2} exhibit generalization performance on eight external validation benchmarks. NuNext achieves the best PQ scores on seven out of eight datasets and the second-best on the remaining one, demonstrating strong cross-domain generalizability. Notably, on GLySAC and CoNSeP, which feature dense nuclei distributions and diverse morphologies, NuNext surpasses all competitors by a clear margin. Fig.~\ref{fig:cmp} presents qualitative comparison results with previous SOTA methods. More comparison results including model size, computational cost and inference efficiency are provided in the appendix. In brief, by leveraging vLLM \cite{kwon2023efficient} with PagedAttention for efficient KV-cache management, our model achieves inference speed comparable to existing methods.

\subsection{Ablation Study}
\begin{table}[t!]
  \begin{minipage}{0.6\textwidth}
    \centering
    \caption{Effect of our proposed modules.}\label{tab:abl}
    \resizebox{0.85\linewidth}{!}{
    \begin{tabular}{ccccccc|c}
    \toprule[1.5pt]
        NSFT & SASS & CoVT & GRPO & LVGF & FGAS & TGR & $F_1$ \\
        \hline
        & & & & & & & 0.587 \\
        \cmark & & & & & & & 0.812 \\
        \cmark & \cmark & & & & & & 0.819 \\ 
        \cmark & \cmark & \cmark & & & & & 0.822 \\ 
        \cmark & & & \cmark & & & & 0.830 \\
        \cmark & \cmark & \cmark & \cmark & & & & 0.835 \\
        \cmark & \cmark & \cmark & \cmark & \cmark & & & 0.837 \\
        \cmark & \cmark & \cmark & \cmark & & \cmark & & 0.838 \\
        \cmark & \cmark & \cmark & \cmark & \cmark & \cmark & & 0.840 \\
        \cmark & \cmark & \cmark & \cmark & \cmark & \cmark & \cmark & 0.842 \\
        \bottomrule[1.5pt]
    \end{tabular}}
  \end{minipage}
  \hfill
  \begin{minipage}{0.39\textwidth}
    \centering
    \includegraphics[width=0.87\linewidth]{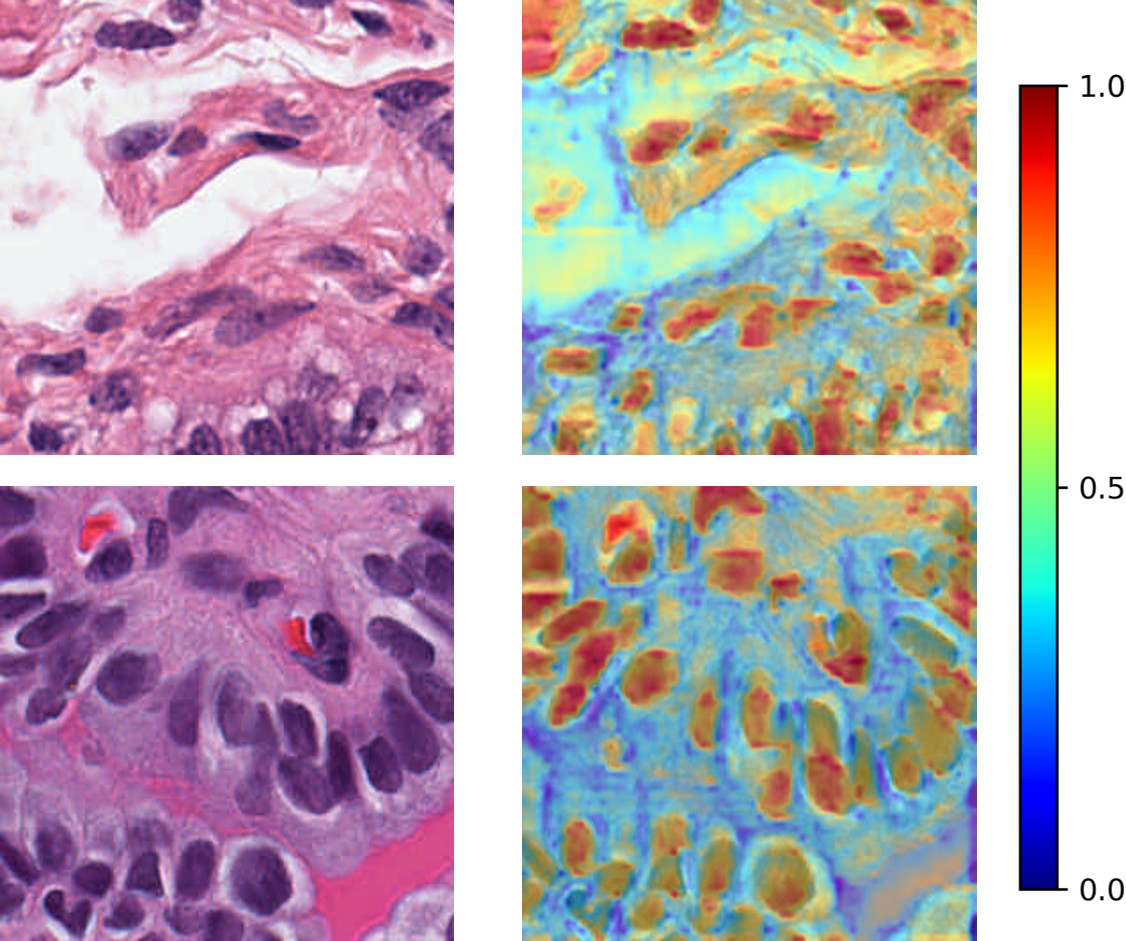}
    \captionof{figure}{Similarity heatmap between latent tokens and visual features. High-response regions align well with nuclei areas.}
    \label{fig:latent}
  \end{minipage}
  \vspace{-10pt}
\end{table}

We investigate the impact of the proposed modules on the detection performance of NuNext on the validation set, including naive supervised fine-tuning (NSFT), spatial-aware soft supervision (SASS), chain-of-visual-thought (CoVT), GRPO with distribution matching reward, low-variance group filtering (LVGF), fine-grained advantage shaping (FGAS) and task-guided reward (TGR). The experimental results in Tab.~\ref{tab:abl} demonstrate that all the proposed modules contribute to the improvement of the model’s performance. Besides, we visualize the cosine similarity between latent token embeddings and visual features in Fig.~\ref{fig:latent}. The heatmaps show high responses in nuclei regions, suggesting that the latent tokens capture the nuclear spatial distribution. This further validates the effectiveness of CoVT. Due to space limitation, more ablation studies such as hyper-paramter analysis are provided in the appendix.

\section{Conclusion}
In this paper, we present NuNext, a generative framework for end-to-end nucleus detection in histopathology images. Unlike previous approaches that rely on complex density map regression, anchor refinement, or query decoding, NuNext directly outputs nucleus centroids via next-point prediction. A series of strategies, including spatial-aware soft supervision and fine-grained advantage shaping, are proposed to boost model performance. Extensive experiments across nine benchmarks demonstrate the superiority of our approach.

\noindent\textbf{Limitation and future work.}
NuNext builds upon a large language model, leading to  considerable storage overhead. We will apply quantization techniques \cite{lin2024awq} to reduce storage requirements while preserving detection performance. For future work, we aim to investigate two directions. (1) \textit{Scaling}: Compared to prior nucleus detection paradigms, our method imposes minimal inductive bias \cite{dosovitskiy2020image}, suggesting that it may benefit more from scaling. We plan to explore the scaling laws of NuNext with respect to data volume and model capacity. (2) \textit{Open-vocabulary extension}: Our architecture naturally supports vision-language interaction, opening the door to open-vocabulary nucleus detection. To this end, we plan to jointly train NuNext on nucleus detection data with precise spatial localization annotations and histology image-caption data that provides rich visual descriptions of nuclear characteristics.

%
%
\bibliographystyle{splncs04}
\bibliography{main}
\end{document}